\documentclass[11pt]{article}

\usepackage[final]{acl}
\usepackage{times}
\usepackage{latexsym}
\usepackage{float}
\usepackage[utf8]{inputenc}
\usepackage[LGR,T1]{fontenc}
\usepackage{textalpha}
\usepackage{CJKutf8}

\usepackage{microtype}

\usepackage{inconsolata}

\usepackage{graphicx}
\usepackage{url}

\usepackage{booktabs}
\usepackage{array}
\usepackage{amsmath}
\usepackage{amssymb}
\newcommand{\lab}[1]{\texttt{#1}}

\title{Fin\_Agent at FinEval 2026: Task-Aware Methods for Multilingual Financial NLP}

\author{
\textbf{Mohotarema Rashid}\quad
\textbf{Tanzina Hossain}\\
\textbf{Xiaoying Song}\quad
\textbf{Lingzi Hong}\\[3pt]
University of North Texas\\[2pt]
{\footnotesize
\{MohotaremaRashid,tanzinahossain,XiaoyingSong\}@my.unt.edu,
Lingzi.Hong@unt.edu
}
}
\begin{document}
\maketitle
\begin{abstract}
We present Fin\_Agent, our submission to the FinEval 2026 shared task on multilingual financial language understanding. Our systems address Greek financial named-entity recognition, Japanese implicit communication recognition, and multilingual financial question answering. For Greek NER, we combine annotation-aware numeric rules with zero-shot Qwen2.5-32B classification, achieving 0.9521 public F1 and 0.8884 private F1. For Japanese implicit communication, an interpretable cue-based ordinal scorer achieves 98.0\% test accuracy. For multilingual financial QA, BGE-M3 retrieval with Qwen3-32B generation achieves 0.3023 and 0.3419 ROUGE-1 F1 on the public and private splits. Key limitations include dataset-specific rules, handcrafted cues, and limited supervision. Implementation code is available 
\url{https://github.com/Rashidnira/finagent-fineval-2026}.
\end{abstract}

\section{Introduction}

Financial markets rely on heterogeneous information from corporate filings, financial statements, earnings announcements, news, and investor-facing communications \citep{rashid2026, rashid2025hong, rashid2025lingzi, zhao2026}. Real-world financial analysis increasingly spans multiple languages and modalities, including reports, news, scanned filings, and meeting recordings \citep{multifinben2026}. Because finance is a high-stakes domain, effective language processing requires both domain knowledge and contextual interpretation \citep{lin2026}. These challenges are amplified in low-resource languages, where financial corpora, expert annotations, benchmarks, and pretrained models remain limited \citep{cao2026, jorgensen2023, plutus2025}. Although multilingual LLMs offer broad cross-lingual capabilities, their financial performance remains uneven across languages, particularly when specialized terminology and complex reasoning are involved \citep{multifinben2026}.

Despite growing interest in applying AI and LLMs to finance \citep{wang2026}, prior work has largely focused on tasks such as financial summarization, question answering, and stock-movement prediction, while multilingual and low-resource financial NLP remains comparatively underexplored \citep{cao2025}. The FinEval 2026 Multilingual Financial Language Understanding shared task addresses this gap through three complementary problems: Subtask 1 focuses on Greek financial NER for extracting structured numeric and textual entities; Subtask 2 targets Japanese implicit communication recognition by modeling the degree of managerial commitment in investor-facing responses; and Subtask 3 evaluates multilingual financial question answering across SEC filings and contemporaneous news in English, Chinese, Japanese, Spanish, and Greek.

\begin{figure}[t]
    \centering
    \setlength{\abovecaptionskip}{4pt}
    \includegraphics[width=\columnwidth]{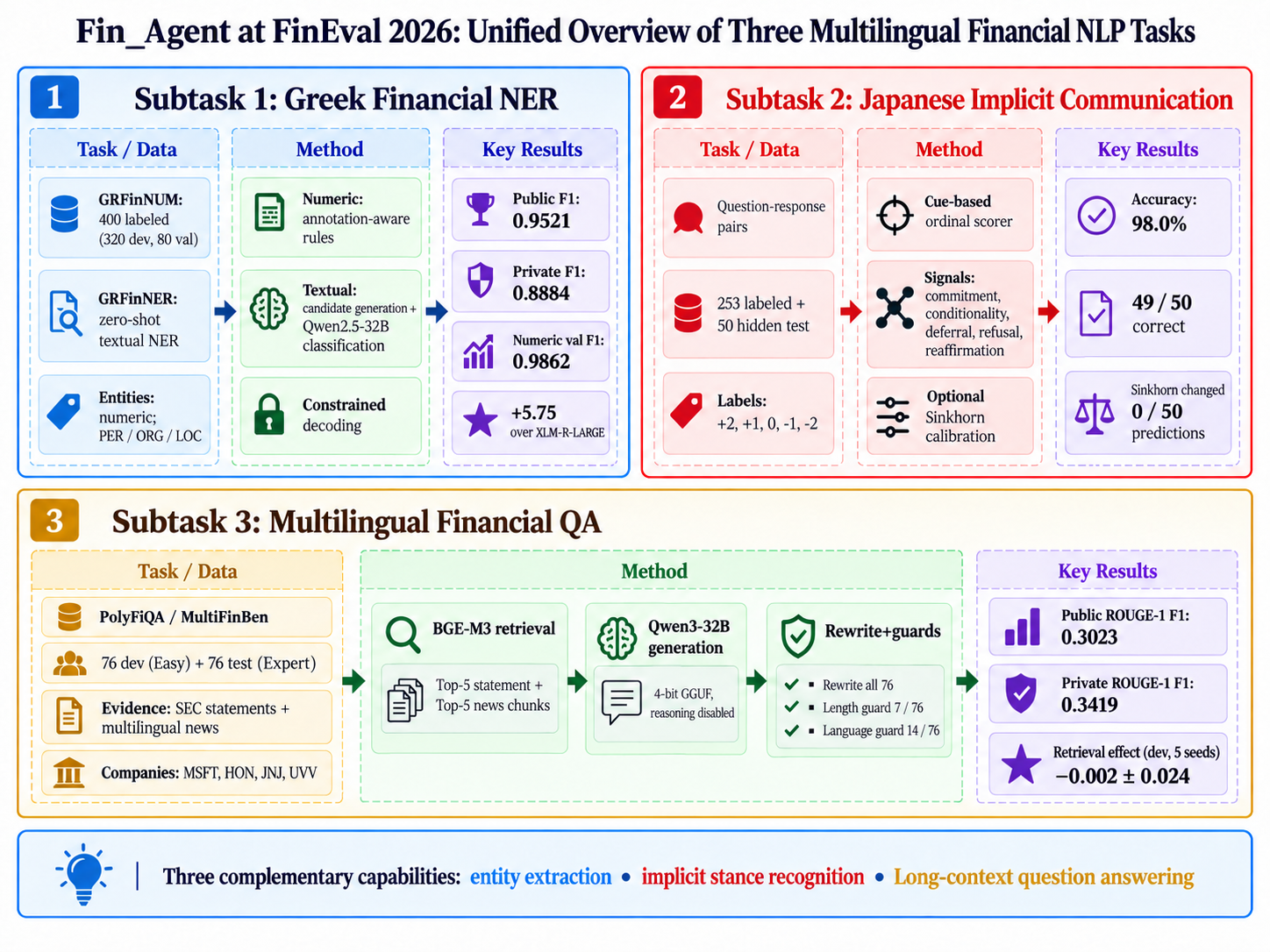}
    \small
    \caption{System architecture.}
    \label{fig:system-overview}
\end{figure}

We develop task-specific systems tailored to the supervision and reasoning
requirements of each subtask (Figure~\ref{fig:system-overview}): a hybrid rule based LLM approach for Greek financial
NER, an interpretable cue-based model for Japanese implicit communication, and
a retrieval-guided generation pipeline for multilingual financial QA.

Our contributions are threefold:
(1) we show that annotation-aware rules can effectively complement LLM-based
classification under limited and asymmetric NER supervision;
(2) we demonstrate that a lightweight, interpretable linguistic model can
accurately capture ordinal commitment in Japanese financial communication
without relying on LLM inference; and
(3) we investigate retrieval-guided and question-level generation for
multilingual financial QA and analyze their effects under controlled settings.
\enlargethispage{2\baselineskip}
 \section{Related Work}
\label{sec:related-work}
 
\subsection{Named-entity recognition in specialized domains}
 
Financial NER extracts structured information from corporate filings,
financial statements, and business documents, with recent work introducing
domain-specific resources and evaluating LLMs for financial entity extraction
\citep{loukas2022finer,luhuo2025}. However, financial NLP resources remain
concentrated in high-resource languages, while domain-specific annotation is
costly and scarce \citep{aguda2024,uthayasooriyar2025}, making non-English
financial NER particularly challenging.

Like other low-resource languages, Greek NER suffers from limited annotated resources and morphological
complexity \citep{lucarelli2007}. Earlier approaches relied on rules and
gazetteers, followed by statistical models and
Greek-specific pretrained models such as GreekBERT
\citep{koutsikakis2020greekbert,michailidis2006greek}. More recently,
\citet{plutus2025} introduced Plutus and showed that Greek financial NER
remains challenging even for modern multilingual and LLM-based models.
To address this gap, Fin\_Agent combines rule-based extraction for structured
financial entities with zero-shot LLM classification for textual entities.
 
\subsection{Implicit communication}
 
Financial communication often conveys managerial intent indirectly through
tone, uncertainty, and strategic wording, providing information beyond
numerical disclosures \citep{bushee2018,larcker2012,li2008,loughran2011liability}.
However, financial NLP has largely focused on sentiment, tone, and deception
rather than implicit managerial commitment \citep{kearney2014}. Japanese
financial NLP has mainly addressed corporate disclosures, financial benchmarks,
and meeting understanding \citep{doi2020tddc,hirano2024,zhu2025mfinmeeting}.
We address this gap using interpretable linguistic cues for commitment,
deferral, reaffirmation, and refusal.

\subsection{Multilingual Financial Question Answering}

Financial question answering requires models to identify and combine numerical
and narrative evidence from corporate disclosures. FinBen broadens evaluation
across financial QA and retrieval-augmented tasks \citep{xie2024}, while
bilingual studies report performance differences between English and Spanish
financial NLP \citep{zhang2024}. However, these settings do not directly test
analyst-style questions requiring joint reasoning over financial statements and
contemporaneous multilingual news. MultiFinBen addresses this gap through
PolyFiQA, which combines SEC statements, news in five languages, and
analyst-style questions \citep{multifinben2026}. Our system uses BGE-M3
retrieval, per-question Qwen3-32B generation, and controlled rewriting to
produce concise, format-compliant answers.

\section{Subtask 1: Greek Financial NER}
\label{sec:s1}
 
\subsection{Task and Data}
\label{sec:s1-task}
 
Subtask 1 focuses on Greek financial NER using two datasets: GRFinNUM for numeric entities and GRFinNER for textual entities. GRFinNUM contains 400 labeled instances (320 development, 80 validation) covering money, percentages, temporal expressions, quantities, and other numeric entities, while GRFinNER targets persons, organizations, and locations and is evaluated in a zero-shot setting because no labeled training examples are provided.
 
\subsection{System}
\label{sec:s1-system}
 
We employ a hybrid architecture reflecting the asymmetric supervision
available across GRFinNUM and {GRFinNER}.

Numeric entities are extracted using ordered regular expressions covering
monetary amounts, percentages, dates, quantities, and other financial
numeric expressions. Ambiguous numerals are classified using nearby
lexical context, and span selection follows the annotation boundaries
observed in the labeled data.
 
For textual NER, we adopt a two-stage candidate-classification approach.
We first generate a high-recall set of candidate spans using capitalization
patterns, corporate legal forms, gazetteers, person-name patterns, acronyms,
and address-like expressions. Each candidate $c_j$ is then classified by
Qwen2.5-32B-Instruct-AWQ \citep{qwen2025technical} as person, organization,
location, or none, using the full passage $x_i$ as context and the label set
$\{\mathrm{PER},\mathrm{ORG},\mathrm{LOC},\mathrm{NONE}\}$. Full rule details are in Appendix~\ref{app:s1-rules}.
The predicted type is
\[
\hat{t}_j =
\arg\max_{t}
p(t \mid x_i,c_j).
\]
Because the model classifies only pre-generated spans rather than freely
generating entities, all accepted predictions remain grounded in the source
passage. A final constrained-decoding stage removes low-confidence, generic,
and overlapping candidates.
 
\subsection{Results and Analysis}
\label{sec:s1-results}

Table~\ref{tab:s1-results} summarizes the main Subtask~1 results.
Our full hybrid system ranked 4th on the public leaderboard with
0.9521 F1 and 6th on the private leaderboard with 0.8884 F1.
Separately, on the numeric validation set, the proposed deterministic
rules achieved 0.9862 F1, outperforming the XLM-R-LARGE baseline
\citep{conneau2020xlmr} by 5.75 F1 points.

\begin{table}[tb]
\centering
\small
\setlength{\tabcolsep}{4pt}
\begin{tabular}{@{}llcc@{}}
\toprule
\textbf{Evaluation} & \textbf{System} & \textbf{F1} & \textbf{Rank} \\
\midrule
Public test  & Hybrid system & 0.9521 & 4 \\
Private test & Hybrid system & 0.8884 & 6 \\
\midrule
Validation & XLM-R-LARGE & 0.9287 & -- \\
Validation & \textbf{Rules (ours)} & \textbf{0.9862} & -- \\
\bottomrule
\end{tabular}
\caption{Subtask~1 official results and numeric-NER ablation.}
\label{tab:s1-results}
\end{table}
\subsection{Limitations and Errors}
\label{sec:s1-limitations}
 
For Subtask~1, Because gold labels for the textual GRFinNER data are unavailable, we assume the main errors involve numeric ambiguity, boundary or type
confusion, and entity omission. Bare numbers could also be difficult to classify when
local context is insufficient, while overlapping spans may lead to incorrect
boundaries or ORG--LOC confusion. Some entities are likely to be missed as the
candidate-generation stage fails to propose the correct span.

\section{Subtask 2: Implicit Communication Recognition}
\label{sec:s2}
 
\subsection{Task and Data}
\label{sec:s2-task}
 
Subtask~2 focuses on identifying the degree of commitment expressed in
Japanese investor-relations responses. Given a question--response pair, the
task is to determine whether the company's response expresses strong
commitment, qualified commitment, neutrality or deferral, weak refusal, or
strong refusal, corresponding to labels $+2$, $+1$, $0$, $-1$, and $-2$,
respectively. The released working data used by our system contain 253 labeled
instances and 50 hidden test instances. Because the target label reflects the
company's communicative stance, our system primarily analyzes the response,
using the question only as supporting context.
 
\subsection{System}
\label{sec:s2-system}

We employ a linguistically motivated scoring framework that maps each
response to five pseudo-logit scores,

\[
\mathbf{s}_i=[s_{+2},s_{+1},s_0,s_{-1},s_{-2}],
\]

using cues for explicit decisions, volition, conditionality, deferral,
soft and hard refusal, reopening, and reaffirmation. Negation is treated
contextually: expressions such as ``there is no change'' are interpreted
as reaffirmation rather than refusal and are masked before hard-negation
detection. The submitted system additionally applies Sinkhorn normalization
\citep{sinkhorn1967,cuturi2013} under an assumed uniform class prior.  The full cue inventory is in Appendix~\ref{app:s2-cues}.
\subsection{Results and Analysis}
\label{sec:s2-results}
 
For Subtask 2, the system achieved 15/15 accuracy on the public split
and 34/35 on the private split, corresponding to a combined accuracy
of 49/50 (98.0\%) across both evaluation splits 
(Table~\ref{tab:s2-official}).

\begin{table}[tb]
\centering\footnotesize
\begin{tabular}{@{}lrrl@{}}
\toprule
Task & Public & Private & Metric \\
\midrule
Subtask 2 Japanese ICR & 100.0\% & 97.14\% & Accuracy \\
\bottomrule
\end{tabular}
\caption{Official Subtask~2 results.}
\label{tab:s2-official}
\end{table}

For Japanese implicit communication, Sinkhorn calibration changes
no test predictions showing that the linguistic scorer rather than
batch-level prior matching determines the final decisions.

\subsection{Limitations and Errors}
\label{sec:s2-limitations}
 
For Japanese implicit communication, errors mainly occur between closely related commitment levels, such as qualified commitment vs. deferral and weak vs. strong refusal. The rule-based cues may also miss unseen or highly context-dependent expressions. 
 
\section{Subtask 3: Multilingual Financial Question Answering}
\label{sec:s3}
 
\subsection{Task and Data}
\label{sec:s3-task}

Subtask 3 evaluates multilingual financial question answering using PolyFiQA from MultiFinBen \citep{multifinben2026}. We use its two datasets: 76 quantitative questions from PolyFiQA-Easy for development and 76 analyst-style questions from PolyFiQA-Expert for evaluation. The evaluation set contains four questions for each of 19 financial-report groups covering MSFT, HON, JNJ, and UVV during 2020–2021. Both datasets use the same reports and evidence contexts but differ in their instructions and question styles. Each question includes SEC financial statements and related news in five languages. The required response contains a concise answer and supporting evidence and cannot exceed 100 words.
\subsection{System}
\label{sec:s3-system}

Because each context is long and combines several sources, we use retrieval to highlight relevant evidence without removing the original context. For each question, BGE-M3 retrieves the five most relevant financial-statement chunks and five most relevant news chunks \citep{chen-etal-2024-m3}. We insert these chunks as an evidence block before the final question. Qwen3-32B then generates one response per question using 4-bit GGUF inference with reasoning disabled \citep{yang2025qwen3}. After generation, a rewrite pass targets 55–65 English words. If the response remains too long, a second pass targets 45 words. If the script check detects characters above U+2000, a language pass rewrites the answer in English with a 70-word target. Final checks flag potentially unsupported numbers and non-verbatim quotations without changing the response. Section~\ref{sec:s3-results} compares the public submission variants and reports a controlled development analysis.

We entered the model-based category and complied with its 70-billion-parameter limit \citep{finnlp2026task}. Qwen3-32B and BGE-M3 contain approximately 33.3B parameters combined and run locally without model API calls; therefore, the OpenRouter requirement for API-based workflows does not apply. Four-bit quantization reduces memory use without changing the parameter count. Appendix~\ref{app:s3-prompts} provide the prompts and processing counts, and Appendix~\ref{app:s3-parameters} reports the parameter accounting.

\subsection{Results and Analysis}
\label{sec:s3-results}

\begin{table}[tb]
\centering\small
\setlength{\tabcolsep}{5pt}
\begin{tabular}{@{}lccc@{}}
\toprule
\textbf{Split} & \textbf{F1} & \textbf{Precision} & \textbf{Recall} \\
\midrule
Public (24)  & 0.3023 & 0.3040 & 0.3528 \\
Private (52) & 0.3419 & 0.3227 & 0.3899 \\
\bottomrule
\end{tabular}
\caption{Official Subtask 3 results.}
\label{tab:s3-results}
\end{table}

\begin{table}[tb]
\centering\small
\setlength{\tabcolsep}{4pt}
\begin{tabular}{@{}lrrrr@{}}
\toprule
\textbf{Configuration} & \textbf{Words} & \textbf{F1} & \textbf{P} & \textbf{R} \\
\midrule
Grouped, no retrieval & 37.6 & .2618 & .344 & .253 \\
Per-question + BGE-M3†  & 91.8 & .2987 & .274 & .386 \\
Per-question + 55-word & 61.2 & .3022 & .321 & .329 \\
Per-question + 72-word & 71.8 & .2944 & .293 & .342 \\
Rewrite + normalization & 69.4 & \textbf{.3023} & .304 & .353 \\
\bottomrule
\end{tabular}

Mean words use all 76 answers; other metrics use the 24 public instances. Final three rows are separate rewrite variants, not cumulative, and share no identical answers. 

† Twenty-three answers exceeded 100 words. We retain this variant to show the performance before rewriting.
\caption{Public-set submission variants.}
\label{tab:s3-ablation}
\end{table}

Table~\ref{tab:s3-results} reports our official results. The final system scores higher on the private split than on the public split, with both precision and recall improving, especially recall. Table~\ref{tab:s3-ablation} records five public submission variants. The largest change occurs when we replace grouped generation without retrieval with question-level generation and retrieval. The three independent rewrite variants differ by less than 0.008 F1. Because they use separate scripts and share no identical answers, these results do not isolate the effects of answer length, language rewriting, or monetary normalization.

Because the largest change introduces retrieval and question-level generation together, the public results cannot isolate either effect. We therefore test all four combinations on PolyFiQA-Easy across five seeds while holding the context preprocessing, instruction format, model, decoding settings, and scorer constant. Mean F1 differs by less than 0.004 across the four configurations. The estimated effects of retrieval, generation granularity, and their interaction are smaller than their variation across seeds. We therefore find no stable advantage for either component under these conditions. Appendix~\ref{app:s3-controlled} reports the complete analysis.

Grouped answers average 33 words and question-level answers average 52 words, compared with a 48-word reference average. Despite this difference, the configurations achieve similar F1 scores. Thus, answer length alone does not explain performance in the controlled analysis.

\subsection{Limitations and Errors}
\label{sec:s3-limitations}
 
Our further error investigation finds six missing evidence labels and flag two numbers and five quotations for review. The two number flags and four quotation flags originate in generation; the length guard adds one quotation flag and all six missing labels (Appendix~\ref{app:s3-errors}). Manual inspection also finds one altered monetary scale, which the validator does not test. Although post-processing introduces some problems, it remains necessary because 23 raw answers exceed the 100-word limit. Hidden references prevent us from measuring the score effect of these cases.
The controlled analysis uses only the development set and may not generalize to the evaluation questions. Repeated evaluation on the small public split may have influenced configuration selection. Finally, our validators flag but do not repair possible errors, and ROUGE-1 measures word overlap rather than factual accuracy.

\section{Conclusion}
\label{sec:conclusion}
 
We present task-aware systems for three complementary multilingual
financial NLP problems. Our results on Greek NER and Japanese implicit
communication show that specialized and interpretable methods can remain
highly effective despite the increasing use of general-purpose LLMs. In
particular, annotation-aware rules substantially outperform the
\textsc{xlm-r-large} baseline for numeric NER, while the Japanese
linguistic scorer achieves $98.0\%$ test accuracy independently of Sinkhorn
calibration. For Subtask 3, our long-context QA pipeline combines retrieval and generation, but the controlled development analysis finds no stable advantage for either retrieval or generation strategy.


\bibliography{custom}

\appendix

% Replace the existing sections "Numeric Rule Inventory for Subtask 1"
% and "Cue Inventory for Subtask 2" with the material below.
% REQUIRED IN THE MAIN PREAMBLE:
% \usepackage{CJKutf8}

\section{Subtask 1: Methods in Detail}
\label{app:s1-rules}

We entered Subtask~1 as a model-based solution.  The only neural model,
Qwen2.5-32B-Instruct-AWQ, has 32B total parameters and is
therefore below the 70B limit.  It was served locally with vLLM 0.6.6.post1.
Numeric NER uses no model; textual NER uses grounded candidate classification.

\subsection{Numeric extraction}

The extractor operates on character offsets and returns ordered
\texttt{(start,end,surface,type)} records.  A character-level claim array
prevents a lower-priority rule from assigning an overlapping span.  Accepted
surfaces are never normalized, and repeated surface--type pairs are retained
because the evaluation treats them as multisets.

The base numeral expression matches digits followed by zero or more period- or comma-delimited digit groups. The exact expression and all ordered rules are available in the accompanying code repository.
Rules are applied in this order: (i) percentages, including ranges;
(ii) three-part dates, decomposed into day, month, and year; (iii) law,
article, paragraph, registry, and accounting-standard identifiers;
(iv) named-month and partial dates, durations, and bare years; and
(v) contextual classification of remaining numerals.  Thus, for example,
\textit{31.12.2020} yields three temporal mentions, while a monetary phrase
normally yields the bare numeral rather than its currency or magnitude word.
This follows the released annotation convention: all labeled money, temporal,
and other-numeric spans are single-token, as are 98\% of percentages.

For a remaining numeral at offsets $[a,b)$, the default context is the 45
characters before $a$ and 40 characters after $b$.  Immediate monetary and
unit tests use narrower windows of 14 characters before and 22 after.
Currency symbols and amount expressions assign \lab{ΧΡΗΜΑΤΑ}; count nouns,
share terminology, parentheses, and physical units assign
\lab{ΠΟΣΟΤΗΤΕΣ}; temporal expressions assign \lab{ΧΡΟΝΙΚΑ}; and an attached
percent sign assigns \lab{ΠΟΣΟΣΤΑ}.  Before fallback classification, the
extractor rejects numbers embedded in Latin names, street addresses, or
hyphenated expressions such as \textit{COVID-19}.  Decimal-like unmatched
numbers default to money when document-level money evidence is present;
otherwise the fallback is quantity. 

On the fixed 320/80 split of the 400 labeled numeric examples, the submitted
rules obtained the following exact entity-level multiset scores.

\begin{table}[H]
\centering\small
\begin{tabular}{@{}lrrr@{}}
\toprule
Split & Precision & Recall & Micro-F1 \\
\midrule
Training (320)   & 0.9877 & 0.9683 & 0.9779 \\
Validation (80) & 0.9882 & 0.9843 & 0.9862 \\
\bottomrule
\end{tabular}
\caption{Subtask~1 numeric-rule performance.}
\label{tab:s1-rule-dev}
\end{table}

\subsection{Textual candidate classification}

Because no labeled GRFinNER examples were supplied, textual NER is zero-shot.
The tokenizer preserves dotted abbreviations, Unicode words, numbers, and
punctuation.  Candidate sources comprise maximal capitalized runs; their
prefixes, suffixes, and contiguous sub-runs; spans ending in Greek or
international legal forms; gazetteer matches; person-shaped bigrams;
title-anchored names; parenthesized acronyms; lowercase geographic heads; and
street-number extensions.  Exhaustive sub-runs are capped at six tokens,
whereas maximal and legal-form spans are capped at sixteen.  Before inference,
we discard candidates shorter than two characters, without letters, longer
than twelve tokens, or beginning with lowercase text.

We built a dictionary of Greek person, organization, and location names from eight existing NER datasets. :
\texttt{wikiann\_el}, \texttt{elNER4}, \texttt{elNER18},
\texttt{openner\_core\_elNER\_ell}, \texttt{greek\_legal\_ner},
\texttt{polyglot\_ner\_el}, \texttt{mapa\_el}, and
\texttt{finerweb\_ell}.  The released code provides the gazetteer-construction procedure and records
the contributing resources.

For each candidate, the model receives the complete passage and candidate span and predicts one label: person, organization, location, or none. We use deterministic, single-token decoding and normalize the returned label log-probabilities. Outputs that do not contain a valid label are treated as none.

For candidate \(c\), we define entity confidence as \(p_{\mathrm{ent}}(c)=1-p(D\mid c)\). Candidates with \(p_{\mathrm{ent}}(c)\geq0.90\) are decoded greedily into non-overlapping spans, prioritizing longer spans and then higher confidence. Each retained candidate is assigned the highest-probability entity class among person, organization, and location.

\section{Subtask 2: Methods in Detail}
\label{app:s2-cues}

\begin{CJK*}{UTF8}{min}

Our method of Subtask~2 is rule based. It
maps the Japanese response to scores in label order
$[+2,+1,0,-1,-2]$. The patterns below give the exact submitted cue
inventory in compact form.

\subsection{Cue inventory}

\begin{table*}[t]
\centering\scriptsize
\setlength{\tabcolsep}{3pt}
\begin{tabular}{@{}p{1.7cm}p{13.8cm}@{}}
\toprule
Family & Submitted Japanese cues (regular-expression alternatives) \\
\midrule
DECIDED & 決議し/いた/済, 決定し/いた/済, 承認し/いた/済, 締結し/済/いた,
確定して/しま/済, 正式に決, 可決, 合意に達し, 署名し, 決まりました,
取締役会で決議/承認/確認/決定 \\
DONE & 完了します/しました/いたします, 開始します, 実施します, 移行します,
適用します, 充当します, 返済します, 継続します, 実行します, 導入します,
払い戻します, 上場します, 着工します \\
NEG\_HARD & いいえ, 予定/計画/こと/事はありません・ございません,
する考えはありません, 意思/方針はありません, しません, いたしません,
行いません, 実施しません, 提供しない, 検討していません, 中止, 終了します,
撤回, 除外して, 白紙, 社内規程, お答えできません, 回答を差し控え,
開示する考えはない, ノーコメント \\
NEG\_SOFT & 難しい, 困難と, 見送る/見送り, 時期尚早, 優先順位は低い,
考えていません, 踏み切る考えは, 慎重に, 控えます, 差し控え \\
REOPEN & ただし/但し, 一方で, 改めて検討/判断/見直し, 再検討, 見直します,
余地は/がある, 状況が変われば, 環境が変わる, 整った段階, 数年後,
将来的には検討, 選択肢の一つ, 高まれば, 柔軟に \\
VOLIT & たいと考え, たいと思う/です, 努めて, 目指し/目指す, 注力して,
進めてまいり, 拡大したい, 引き上げたい, 検討してまいり, 検討を進め,
前向き, 想定している, 判断します, 排除しておらず \\
COND & れば, 次第, 整えば, 得られれば, できれば, 想定どおり, 条件が,
確認できれば, 進捗次第, 市況次第, 場合には, 可能性があります, 応じて \\
NEUTRAL & 検証して, 精査して, 分析して, 策定過程/策定中, 評価している,
確認している, 段階です, 据え置き, 次回の説明, 発表時に説明,
お示しします, 公表します, 双方, 総合的に勘案/判断, 継続的に評価,
検討している段階 \\
NOPROMISE & 確約はできない, 約束は/できない, 保証は/できない,
決まっておりません, 確定しておりません, 決定した事実はありません,
決定しておりません \\
REAFFIRM & 変更は/もありません・ございません, 変更はない/無い,
変更せず, 変更しません, 計画どおり継続, 従来どおり変更なし \\
NUM/DATE & 数字 followed by 億円, 百万円, 万株, 株, 円, \%, 拠点, 店, 名,
件, 年, 月, 日; 20XX年, 月日, 今期末, 来年度/下期, 本日, 来月, 次期,
期限まで \\
\bottomrule
\end{tabular}
\caption{Compact rendering of the exact Subtask~2 regex inventory.  Slash-separated
forms denote alternatives; the released source preserves the precise grouping and suffixes.}
\label{tab:s2-full-cues}
\end{table*}

REAFFIRM spans are removed before counting hard negation.  This prevents
\textit{変更はありません} (``there is no change'') from being treated as a
refusal while retaining a separate genuine refusal in the same response.
Bare \textit{ありません} is deliberately absent from NEG\_HARD because it
often negates a fact rather than the proposed action.

\subsection{Scoring function}

Let $c_f$ be the number of matches from family $f$ and
$m_k(c)=\min(c,k)$.  Scores begin at zero.  Define
\begin{align}
d={}&2.0m_2(c_{\mathrm{DECIDED}})+0.8m_2(c_{\mathrm{DONE}}),\\
u={}&0.7m_2(c_{\mathrm{NUM}})+0.7m_2(c_{\mathrm{DATE}}).
\end{align}
If $d>0$, add $d+u$ to $s_{+2}$ and $0.5d$ to $s_{+1}$; under hard
negation, instead add $d+u$ to $s_{-2}$ and $0.4d$ to $s_{-1}$.
Hard negation adds $2.2m_2(c_{\mathrm{NEG\_HARD}})$ to $s_{-2}$ and
$0.9m_2(c_{\mathrm{NEG\_HARD}})$ to $s_{-1}$.  Soft negation adds
$2.0m_2(c_{\mathrm{NEG\_SOFT}})$ to $s_{-1}$ and
$0.7m_2(c_{\mathrm{NEG\_SOFT}})$ to $s_{-2}$.  Negation with REOPEN adds
1.8 to $s_{-1}$ and subtracts 1.4 from $s_{-2}$; without REOPEN it adds
0.9 to $s_{-2}$.

VOLIT and COND add
\begin{equation}
w\bigl[1.5m_2(c_{\mathrm{VOLIT}})+0.9m_2(c_{\mathrm{COND}})\bigr]
\end{equation}
to $s_{+1}$, where $w=0.5$ under hard negation and 1 otherwise.
NOPROMISE adds 1.0 to $s_{+1}$ and subtracts 1.5 from $s_{+2}$.
VOLIT or COND without $d$ subtracts 1.0 from $s_{+2}$.  REAFFIRM without
hard negation adds 1.2 to $s_{+2}$ and 0.6 to $s_0$.  NEUTRAL adds
$1.6m_3(c_{\mathrm{NEUTRAL}})$ to $s_0$; a response with no negation,
VOLIT, or decision adds another 1.5 to $s_0$, whereas any negation subtracts
1.2.  No polarity reversal is applied when the question proposes a negative
action.  NumPy \texttt{argmax} breaks ties in label order
$[+2,+1,0,-1,-2]$.

\subsection{Split, calibration, and results}

The training set contains 253 labeled examples, distributed as
$+2$: 29, $+1$: 139, $0$: 72, $-1$: 11, and $-2$: 2. We construct a
fixed 26-example development split by sampling six examples from each
class, except $-2$, for which both available examples are used.

The pipeline applies 80 log-space Sinkhorn iterations with a uniform target prior, $\boldsymbol{\pi}=(0.2,0.2,0.2,0.2,0.2)$,
temperature 1.0, and strength 1.0. Final labels are selected using
row-wise argmax. Sinkhorn changes none of the 50 evaluation predictions,
indicating that the final predictions are determined by the rule-based
scorer. As Table~\ref{tab:s2-dev} shows, the likely test-set improvement reflects
shorter, clearer test responses with more explicit decision and refusal
cues, rather than a benefit from Sinkhorn calibration. Since the evaluation labels and their distribution are unavailable, we cannot conclusively determine which dataset characteristics produced the large development–evaluation difference.

\begin{table}[t]
\centering
\small
\begin{tabular}{lcc}
\toprule
Decision rule & Accuracy & Macro-recall \\
\midrule
Raw rule-score argmax & 0.3846 & 0.4000 \\
Sinkhorn, strength 1.0 & 0.2692 & 0.3000 \\
\bottomrule
\end{tabular}
\caption{Subtask 2 results on the fixed 26-item development split.}
\label{tab:s2-dev}
\end{table}

\end{CJK*}

\section{Subtask 3: Methods in Detail}
\label{app:s3-methods}

\subsection{Prompts and Processing Counts}
\label{app:s3-prompts}

The submitted pipeline uses one generation stage followed by an initial rewrite and two conditional guards. Every model call uses Qwen3-32B with 4-bit GGUF inference and reasoning disabled. We use temperature 0.7, top-p 0.8, top-k 20, a 4,096-token output limit, and a 32,768-token context window. BGE-M3 retrieves the five highest-ranked financial-statement chunks and five highest-ranked news chunks for each question. The headings below report each stage’s trigger and processing count.

\newenvironment{prompt}%
  {\par\vspace{2pt}\begingroup\footnotesize\leftskip=1em\noindent\ignorespaces}%
  {\par\endgroup\vspace{2pt}}

\paragraph{Evidence block (all 76).} We insert the retrieved chunks into the organisers’ query immediately before its final Question: line without replacing the original context.
\begin{prompt}
Highlighted Evidence (automatically retrieved, verbatim excerpts from the
context above; check these first, but the full context above remains
authoritative): [Financial statement excerpts] \dots{} [News excerpts]
\end{prompt}

\paragraph{Rewrite (all 76).} Applied to every generated answer.
\begin{prompt}
You rewrite financial-QA answers to a strict house style. Rewrite the
answer below. Rules:
\begin{list}{---}{\leftmargin=1em\itemindent=0em\labelwidth=0.6em
  \labelsep=0.4em\topsep=1pt\parsep=0pt\itemsep=1pt}
\item Target 55--65 words total, NEVER more than 75.
\item ENGLISH ONLY. If the answer contains any Chinese, Japanese, Greek, or
Spanish text (including quotes), replace it with a faithful English
rendering of the same information. No non-Latin characters may remain.
\item Keep the exact labeled structure of the original (e.g.\ ``Answer:
\dots'' and the ``\dots{} Evidence: \dots'' line with its exact label).
\item Write monetary amounts compactly: \$27.2B, \$416M, \$3.5M --- but ONLY
when the magnitude is clear from the original wording (e.g.\ ``US\$14,014
million'' $\rightarrow$ ``\$14.0B''). If the scale of a raw number is not
explicit, keep it unchanged. Never alter what a number means. Keep
percentages as printed (e.g.\ 18.6\%).
\item Keep every distinct fact. Reuse the question's key terms in the
wording.
\item Remove filler, hedging, repetition, and meta-commentary.
\item Output ONLY the rewritten answer text, nothing else.
\end{list}
\end{prompt}

\paragraph{Length guard (7 of 76).} Applied only when the rewritten answer
still exceeded 100 words.
\begin{prompt}
Rewrite the answer below to AT MOST 45 words total. Keep the exact labeled
structure (Answer: / \dots{} Evidence: lines), keep every figure exactly as
written, keep at most ONE short verbatim quote (drop the rest; never alter
or translate inside quotation marks). Output ONLY the rewritten answer.
\end{prompt}

\paragraph{Language guard (14 of 76).} Applied when the script detector found characters above U+2000 after the rewrite and length-guard stages. This check detects Chinese and Japanese characters and some typographic punctuation, but not Greek, Cyrillic, or Spanish written in the Latin script.
\begin{prompt}
The answer below contains non-English text (Chinese, Japanese, Greek, or
Spanish). Rewrite it in ENGLISH ONLY, at most 70 words: translate the
non-English parts faithfully into English prose (no quotation marks around
translations), keep the labeled structure (Answer: / \dots{} Evidence:
lines), keep all figures exactly. Not a single non-Latin character may
remain. Output ONLY the rewritten answer.
\end{prompt}

Of the 76 generated answers, 23 exceeded the 100-word limit before rewriting, with the longest containing 153 words. Seven still exceeded the limit after the initial rewrite and therefore received the length guard.

\subsection{Parameter Accounting}
\label{app:s3-parameters}

The shared task defines two participation categories. The published participation requirements state that “Participants may submit either a model-based solution or an API-based agent/workflow solution.” For model-based solutions, “The submitted model must not exceed 70 billion total parameters (70B).” For API-based agent and workflow solutions, “All model API calls must be made through OpenRouter,” and “Each model used in the workflow must not exceed 70 billion total parameters (70B).” For Mixture-of-Experts models, “Eligibility is determined by the model’s total parameter count, not the number of parameters activated during inference” (\citep{finnlp2026task}). We entered the model-based category, used the released weights without further training, and made no model API calls. The OpenRouter requirement is scoped to the API-based category and does not apply to our submission. Table~\ref{tab:s3-parameters} includes every learned component. We conservatively sum the generator and retriever rather than counting the generation model alone.

\begin{table}[H]
\centering\small
\begin{tabular}{@{}lrr@{}}
\toprule
Component & Role & Parameters \\
\midrule
Qwen3-32B & Generation/rewriting & 32,762,123,264 \\
BGE-M3 & Passage retrieval & 567,754,752 \\
Deterministic code & Chunking/validation & 0 \\
\midrule
Total & & 33,329,878,016 \\
\bottomrule
\end{tabular}
\caption{Learned components and total parameter count for Subtask~3.}
\label{tab:s3-parameters}
\end{table}

We serve Qwen3-32B using 4-bit GGUF weights. Quantization reduces storage precision without removing parameters. Both models use open weights and run on local hardware without contacting a hosted inference service. BGE-M3 generates no text; it ranks source passages that are inserted into the generation prompt. Qwen3-32B performs all answer generation and rewriting, while the chunker and validation checks use no learned parameters. 
The parameter counts come from the released model artefacts. The
Qwen3-32B count is taken from the
\texttt{general.parameter\_count} field of the served model file, and
the BGE-M3 count is calculated from the tensor shapes in its released
weight file.

\subsection{Controlled Development Analysis}
\label{app:s3-controlled}

Table~\ref{tab:s3-ablation} compares variants that differ in more than one respect at a time. To separate the two design choices, we built a grid of four configurations and ran each on the 76 development instances with five seeds.

\begin{table}[H]
\centering\footnotesize
\setlength{\tabcolsep}{3pt}
\begin{tabular}{@{}lll@{}}
\toprule
Label & Generation & Retrieval \\
\midrule
G   & Grouped (4 questions/call) & None \\
G+R & Grouped                    & Per-question evidence \\
Q   & One call per question      & None \\
Q+R & One call per question      & Evidence before question \\
\bottomrule
\end{tabular}
\caption{Controlled-analysis configurations.}
\label{tab:s3-controlled-design}
\end{table}

The cells differ only in retrieval and generation strategy. The generation strategy includes question grouping, the number of model calls, and output allocation. The 4,096-token output limit never binds: across the 190 grouped calls the longest completion was 694 tokens and every call ended normally.
We built all four prompt sets from a single source. Each retrieval prompt records where its evidence block was inserted, so removing that span recovers the original query exactly, and the same span is what we re-insert for the retrieval cells. Every cell uses the same raw context, the same model and quantization, the same decoding settings, the same output limit, and the same scoring code. We apply no rewriting or normalization. One difference cannot be removed: answering four questions in one call requires an instruction stating how to separate the answers. That instruction is identical in G and G+R, so it cannot affect the retrieval comparison.
We ran this analysis after the evaluation phase closed, with a 40,960-token context window and the Vulkan backend of our inference server on an H100 GPU, sampling at temperature 0.7 with seeds 1 to 5. This environment differs from the one that produced our submission, so the grid is a controlled analysis rather than a reproduction of the submitted runs. Using one environment for all four cells supports internally consistent comparisons.

\begin{table}[H]
\centering\small
\begin{tabular}{@{}lcc@{}}
\toprule
Configuration & Default F1 & Multiscript F1 \\
\midrule
G   & $0.4547\pm0.0151$ & $0.4396\pm0.0191$ \\
G+R & $0.4525\pm0.0103$ & $0.4391\pm0.0097$ \\
Q   & $0.4564\pm0.0054$ & $0.4383\pm0.0087$ \\
Q+R & $0.4559\pm0.0080$ & $0.4406\pm0.0092$ \\
\bottomrule
\end{tabular}
\caption{Development ROUGE-1 F1, mean and standard deviation over five seeds.}
\label{tab:s3-controlled-scores}
\end{table}

We compute each effect within a seed and then summarise the five differences, which accounts for seed-level variation shared across the four configurations.

\begin{table}[H]
\centering\footnotesize
\setlength{\tabcolsep}{3pt}
\begin{tabular}{@{}lrr@{}}
\toprule
Contrast & Default & Multiscript \\
\midrule
G+R$-$G       & $-0.0022\pm0.0239$ & $-0.0005\pm0.0269$ \\
Q+R$-$Q       & $-0.0005\pm0.0072$ & $+0.0023\pm0.0141$ \\
Q$-$G         & $+0.0017\pm0.0195$ & $-0.0013\pm0.0231$ \\
(Q+R)$-$(G+R) & $+0.0034\pm0.0113$ & $+0.0015\pm0.0110$ \\
Interaction   & $+0.0017\pm0.0250$ & $+0.0028\pm0.0273$ \\
\bottomrule
\end{tabular}
\caption{Within-seed paired effects on development ROUGE-1 F1 (mean $\pm$
s.d.\ over five seeds).}
\label{tab:s3-controlled-effects}
\end{table}

Every estimate is smaller than the variation across the five runs, and the four configurations fall within 0.004 F1 of one another. With five seeds these figures describe the spread we observed; they are not a significance test. A small estimate means an effect we could not resolve at this sample size, not the absence of one.

The largest descriptive differences appear for R\&D and cash-flow questions, but the patterns vary with retrieval. Each question type contains only 19 instances, so we draw no general conclusions.

We release the prompt templates and scripts required to generate the four
configurations and reproduce the controlled analysis. A consistency audit checks that the cells differ only as described, that every run produced 76 answers, that all generations used the pinned model at the recorded seeds and window, and that replaying a cell from cache reproduces its answers exactly.

\begin{table}[H]
\centering\small
\begin{tabular}{@{}lrr@{}}
\toprule
Configuration & Mean words & Ratio to reference \\
\midrule
Reference & 47.9 & 1.00 \\
G & 33.3 & 0.70 \\
G+R & 35.3 & 0.74 \\
Q & 51.6 & 1.08 \\
Q+R & 55.8 & 1.16 \\
\bottomrule
\end{tabular}
\caption{Mean answer length in the controlled development analysis.}
\label{tab:s3-controlled-length}
\end{table}
\begin{table}[H]
\centering\footnotesize
\begin{tabular}{@{}lrrrr@{}}
\toprule
Configuration & Revenue & Balance sheet & Cash flow & R\&D \\
\midrule
G & .422 & .460 & .312 & .625 \\
G+R & .420 & .464 & .309 & .617 \\
Q & .424 & .483 & .340 & .578 \\
Q+R & .442 & .482 & .308 & .591 \\
\bottomrule
\end{tabular}
\caption{Development F1 by question type, default scorer, averaged over five
seeds. Each type contains 19 instances.}
\label{tab:s3-controlled-type}
\end{table}

\subsection{Sources of Remaining Validator Flags}
\label{app:s3-errors}
Our pipeline applies three post-processing stages after generation. To find out whether the flags raised on our submitted answers come from the model or from our own processing, we reproduced each stage from the cached responses and applied the same reference-free validator after each one. The final stage reproduces the submitted file byte for byte, so the trace describes the system we submitted.

The three classes of validator flags arise at different stages. Untraceable-number flags originate in generation, and the rewrite clears three of the five affected rows without adding any. Four quotation flags originate in generation, while the length guard adds one. The length guard introduces all six missing evidence labels, dropping the label from six answers while compressing them below the word limit.
That last result is a failure to follow an explicit instruction rather than a missing one. The length-guard prompt in \ref{app:s3-prompts} requires the labeled structure to be kept, and the model still dropped the label in six of the seven answers the stage touched. We conclude that a required output element needs a deterministic check after generation, and that a validator which only flags such cases, as ours does, will not prevent them.

\begin{table}[H]
\centering\footnotesize
\setlength{\tabcolsep}{3pt}
\begin{tabular}{@{}lrrrr@{}}
\toprule
Flagged rows & Raw & +Rewr. & +Len. & +Lang. \\
\midrule
Untraceable numbers    & 5  & 2 & 2  & 2 \\
Non-verbatim quotes    & 4  & 4 & 5  & 5 \\
Missing evidence label & 0  & 0 & 6  & 6 \\
Rows with any flag     & 19 & 8 & 11 & 11 \\
\bottomrule
\end{tabular}
\caption{Validator flags after each stage. Values are numbers of rows. The final row reports the union of all flagged rows, including checks not listed separately, chiefly answers longer than 110 words at raw generation. Because one answer may carry multiple flags, this value need not equal the sum of the three displayed classes. The label check tests for an exact string and is reliable; the number and quotation checks are heuristic and may flag derived figures or normalised wording.}
\label{tab:s3-validator}
\end{table}

\begin{table}[H]
\centering
\small
\setlength{\tabcolsep}{3pt}
\begin{tabular}{@{}lrrrr@{}}
\toprule
Stage & Mean words & Max words & Over 100 \\
\midrule
Raw generation   & 91.8 & 153 & 23 \\
+ Rewrite        & 73.7 & 126 &  7 \\
+ Length guard   & 69.4 &  99 &  0 \\
+ Language guard & 69.4 &  99 &  0 \\
\bottomrule
\end{tabular}
\caption{Answer length after each stage of the submitted post-processing pipeline. The two compression variants in
Table~\ref{tab:s3-ablation} were produced by separate scripts and are not sequential stages
of this pipeline.}
\label{tab:s3-stage-length}
\end{table}

After the language guard, 10 answers still triggered the code-point check because they contained curly quotation marks or dashes, not Chinese or Japanese text.

\end{document}